# Towards an environmental balance for nature

Jiali Li[1], Guowangchen Liu[2,3], Aiqing Zhu[4], Dongfeng Li[2], Chengzhi Hu[1], Jiuhui Qu[1*]

[1]Key Laboratory of Environmental Aquatic Chemistry, State Key Laboratory of Regional Environment and Sustainability, Research Center for Eco-Environmental Sciences, Chinese Academy of Sciences, Beijing 100085, China
[2]Key Laboratory for Water and Sediment Sciences, Ministry of Education, College of Environmental Sciences and Engineering, Peking University, Beijing 100871, China
[3]School of Environmental Science and Engineering, Suzhou University of Science and Technology, Suzhou 215009, China
[4]Department of Mathematics, National University of Singapore, 10 Lower Kent Ridge Road, 119076, Singapore

## Abstract

Human activities are reshaping environmental systems by altering the flows and transformations of matter and energy. Steering these changes towards sustainability requires more than defining environmental thresholds and tracking indicators. It requires understanding which desirable environmental states can persist, what processes and relations sustain them and through which pathways systems can reach these states. Here we propose a computable first-principles state–path framework for environmental balance, linking natural processes and human activities to the emergence and dynamic maintenance of environmental states. By integrating physical principles with environmental data and artificial intelligence, this framework could make environmental states, their underlying balance relations and the pathways connecting them quantitatively tractable. Building it will require a foundation theory for understanding human–nature relations, charting pathways towards sustainable states and translating environmental goals into the technological capabilities needed to achieve them.

## Introduction

Humanity is well into the Anthropocene, in which human activities have become a dominant force reshaping the Earth system[1–3]. At root, human activities alter energy flows and matter inputs, including pollutants and greenhouse gases, creating imbalances that reshape biogeochemical cycles and ecosystem structure. The accumulation of $CO_2$ and other greenhouse gases, for example, alters Earth's radiative balance and drives accelerating climate change[4,5]. The past 11 years have been the warmest on record, current policies still point to about 2.8 °C of warming this centur. Emerging contaminants are becoming more widespread in air-water-soil systems, and six of the nine planetary boundaries have already been transgressed[6–9]. These problems persist, because human activities have reorganized the states of environmental systems faster than we have come to understand them[10]. Knowing which desirable environmental states can persist under natural and human influence, and along which pathways systems can reach them is crucial for improving understandings of environmental systems. This perspective proposes a computable first-principles state–path framework for environmental balance as a route to such an understanding, and outlines the theories needed to support it.

Environmental change is commonly tracked through hydrometeorological, chemical concentration, and ecological indicators[11,12]. Global observing systems, from satellite remote sensing to in situ monitoring networks and long-term ecological records, now deliver such measurements at unprecedented resolution and coverage. These measurements are indispensable, but they describe a system without explaining how its state is generated. At a fundamental level, environmental states change through exchanges and transformations of matter and energy, organized by internal structure and feedbacks (Fig. 1a). Similar indicator values can therefore be maintained by different internal relations and carry different implications for stability, recovery and maintenance[13]. Improvement in a single indicator may also fail to place the target system on a more maintainable trajectory if the relations governing it remain unchanged. The resulting theoretical challenge is to infer the quantitative matter–energy relations from observed indicators' dynamics. These relations generate and sustain environmental states and govern how they can evolve.

Existing frameworks address parts of this challenge. Planetary boundaries identify conditions under which Earth-system risk increases[14,15]. Resilience theory examines whether systems persist or change regime, while adaptive governance focuses on decisions under uncertainty[16,17]. These frameworks answer different questions, and their analytical objects are not ordinarily represented within a shared state-path structure. What remains missing is a transferable and computable framework that explains which states can persist under given boundaries, which relations maintain them and how changing boundaries or mechanisms reshape the pathways between states.

We propose an environmental balance framework as a step towards filling this gap. It represents an open environmental system through its joint matter-energy state, organized by internal relations, conditioned by natural and human boundaries, and transformed along continuous pathways. Rather than adding another aggregate indicator, it offers a set of connected objects through which physical accounting, stability analysis and environmental governance can be related. The following sections define the matter–energy state, environmental free energy (EFE) and balance domains; distinguish the domain families generated under different boundary conditions; introduce hypotheses H1–H4 and a route towards computation; and consider how desirable environmental states can be selected and reached.

## Environmental balance

To explain how an environmental system changes, we must first specify what constitutes its state. Hydrometeorological variables, chemical concentrations and ecological indicators describe different aspects of that state, but their values arise from the organization and transformation of matter and energy (Fig. 1a). The environmental balance framework therefore represents the state at a chosen scale through the matter- and energy-related variables needed to distinguish relevant system configurations, $X = [M, E]$. $M$ comprises matter state variables such as quantity, composition, chemical form and spatial distribution. $E$ comprises energy-related state variables such as temperature, radiation fields and the thermal, chemical, kinetic and potential energy associated with the matter state. The composition of $X$ depends on the target system, question and scale. At coarser scales, these variables may represent the collective effects of finer-scale states, allowing system-level paths to remain traceable to the finer-scale mechanisms that generate them.

Once $X$ has been defined, each observation locates the system at a point in matter–energy state space, and successive observations trace a path $X(t)$. This path records how the state changes, but does not explain why some regions are repeatedly occupied, why a displaced system returns or moves elsewhere, or how changing boundaries open and close possible routes. These are the central questions of a state–path theory. Answering them requires a representation that relates observed trajectories to the stable structure of state space and to the conditions that reshape it. Thermodynamics addresses an analogous problem through scalar potentials, including Gibbs and Helmholtz free energies, that order states under specified constraints[18].

Non-equilibrium landscape approaches retain this organizing role for systems sustained by fluxes[19–23]. The environmental balance framework adapts this state-space logic to environmental systems by defining environmental free energy (EFE) as a learnable scalar function of $X$. Conditioned on internal organization $\theta$, boundary conditions $\lambda$ and observational scale $\tau$, $EFE$ organizes the coupled matter–energy configuration into a landscape of stable structure (Fig. 1b). This landscape locates balance domains, differentiates their relative stability and identifies possible connections between them.

EFE becomes relevant to dynamics because an environmental system can continue to change even when exchanges across its chosen boundary are temporarily removed. Matter and energy already within the boundary continue to redistribute and transform. If external loading and energy to a lake ceased, for example, heat would still redistribute, organic matter would decompose and nutrients would continue to cycle to the end[24]. We call the change that remains the intrinsic dynamics of the boundary-closed system, $D_{self}$. The EFE landscape represents the stable structure within which these internal processes unfold. Its gradient shows how EFE changes between neighboring states, while the internal physical, chemical and biological relations represented by $\theta$ convert those differences into actual changes in $X$.

Exchange with surrounding natural systems, human influence and randomness can then be added to the intrinsic dynamics:

$$\frac{dX}{dt} = \underbrace{D_{self}\left(\nabla_X EFE;\theta\right)}_{intrinsic\,dynamics} + \underbrace{U_N(t;X)}_{natural\,exchange} + \underbrace{U_H(t)}_{human\,influence} + \underbrace{\xi(t)}_{randomness}$$

Here, $U_N$ denotes natural matter–energy exchange whose magnitude and direction may depend on the current system state $X$, $U_H$ denotes direct human influence and $\xi(t)$ denotes unresolved variability. Natural processes and human actions may also alter $\lambda$ or $\theta$, reshaping the EFE landscape and intrinsic dynamics. For the lake introduced above, inflowing streams, precipitation and solar radiation constitute natural exchange, continuously delivering matter and energy across its boundary. Sewage discharge, aquaculture and similar human activities enter as direct human influence, adding matter–energy inputs that may reinforce, offset or redirect the natural fluxes.

Taken together, these terms determine the system's instantaneous evolution. Whether the system remains balanced, however, depends on how their combined effects accumulate over the selected timescale. Mismatch arises when these effects are not compensated within the system. Such mismatch may be temporary, producing fluctuations or recoverable displacement. Imbalance develops only when mismatch accumulates directionally and drives sustained change in $X$ (Fig. 1c).

Because such fluctuations and recoverable displacements can occur without loss of balance, environmental balance is represented as a domain rather than a fixed point. For a given system boundary, external driving and observational scale, a balance domain is a bounded region of state space within which long-term trajectories remain statistically confined (Fig. 1c). Directional accumulation of mismatch can instead drive the system out of its current domain and towards another (Fig. 1c). If this accumulation is accompanied by changes in θ or λ, the EFE landscape itself is reshaped, changing both the domains and the paths between them. Environmental balance therefore denotes dynamic maintainability rather than stasis or thermodynamic equilibrium. Taken together, this formulation provides a basis for identifying which environmental states can persist, how intrinsic dynamics, natural exchange and human influence jointly sustain them and under what conditions they change.

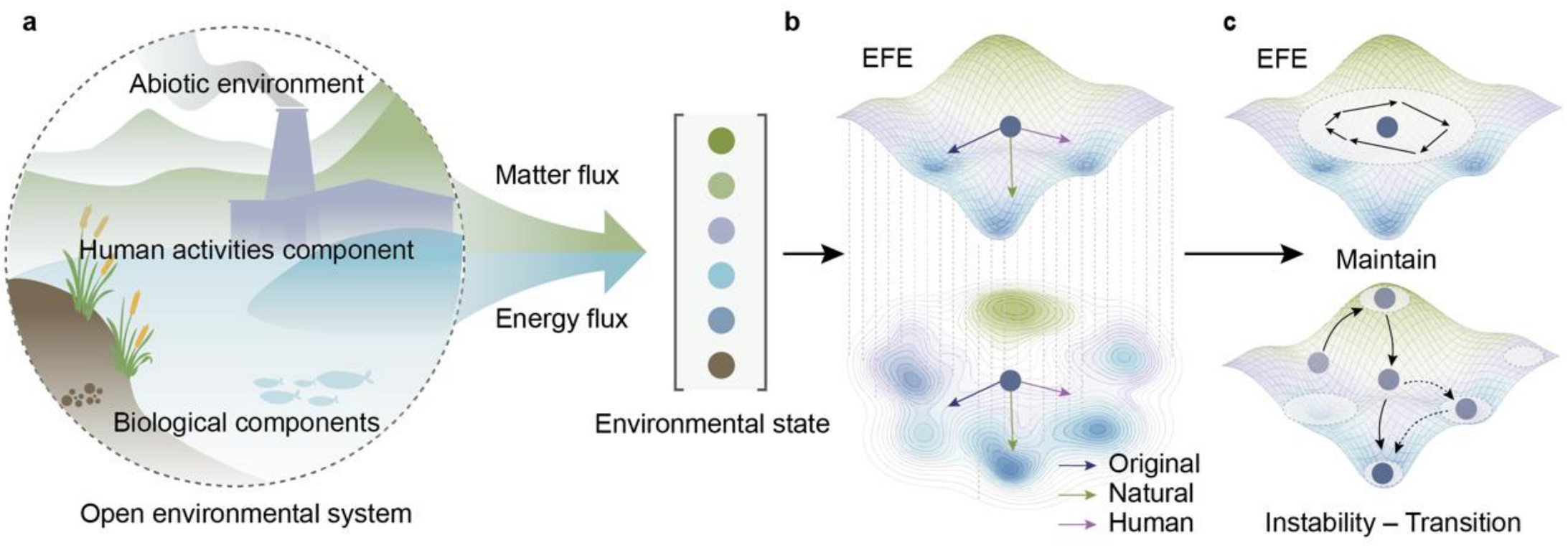

**Fig. 1 | Environmental balance as a state maintained by the organization of matter and energy. a**, Matter and energy fluxes generate and alter environmental state $X$. **b**, EFE organizes states into a landscape of balance domains and possible paths. Blue, green and purple arrows denote intrinsic dynamics, natural exchange and human influence, respectively. **c**, Bounded mismatch confines trajectories within a balance domain (top), whereas directional accumulation drives transitions between domains (bottom). Dashed outlines mark domain boundaries; circles represent environmental states.

## Boundary-conditioned balance domains

These full dynamics above can then be analyzed to determine how each contribution changes the states the system can maintain. At the selected observational scale $\tau$, each balance domain is defined as a connected, dynamically coherent region of admissible state space associated with a locally attracting stationary or quasi-stationary distribution. It represents a distinct stable or metastable organization of the system and cannot be further divided into separate locally attracting regions at that scale. The domains considered below retain this common definition but are obtained by conditioning the full dynamics on different combinations of intrinsic dynamics, natural exchange and human influence. This distinction matters because similar states may depend on different combinations of these contributions and may therefore differ in vulnerability, maintenance requirements and responses to intervention. The framework separates these contributions by evaluating the same target system under systematically varied boundary conditions. The resulting intrinsic, naturally conditioned and human-conditioned balance domains are conditional descriptions of the same system rather than successive stages of environmental change. Their comparison reveals what the system can sustain internally, what depends on surrounding natural systems and how human action changes the states that can persist.

When natural exchange and human influence are removed, ($U_N$=$U_H$=0), the system evolves only through $D_{self}$. All admissible states satisfying $\nabla_X EFE = 0$ and having a positive-definite Hessian, $\nabla_X^2 EFE > 0$, constitute the intrinsic balance states. At the observational scale $\tau$, these states and their surrounding fluctuations form one or more intrinsic balance domains, collectively denoted $\beta_0$. These domains provide a theoretical boundary-closed reference for what the system can maintain through its internal organization using the matter and energy already within its boundary (Fig. 2a).

Restoring natural exchange while excluding human influence opens the system boundary and generates naturally conditioned balance domains, collectively denoted $\beta_N$. Because $U_N$ continuously acts on $X$, these domains must lie where $\nabla_X EFE \neq 0$. At zero gradient, $D_{self}$ would provide no intrinsic response to compensate for natural exchange, and $X$ would move persistently in the direction imposed by $U_N$. Within $\beta_N$, gradient-driven intrinsic dynamics compensate for the effects of $U_N$ over the selected timescale, keeping cumulative mismatch bounded even as matter and energy fluxes continue through the system. Changes in natural exchange may shift, enlarge, contract, create or eliminate balance domains. Comparing $\beta_N$ with $\beta_0$ therefore reveals how natural exchange reshapes the system's maintainable state space relative to its boundary-closed condition (Fig. 2b).

Introducing human influence adds $U_H$ to the open-system dynamics and generates human-conditioned balance domains, collectively denoted $\beta_H$. Depending on its direction, $U_H$ may reinforce, counteract or redirect $U_N$, changing the EFE gradient required for dynamic compensation. The resulting $\beta_H$ captures whether human action shifts the balance supported by natural exchange or enables a different domain to persist. Comparing $\beta_H$ with $\beta_N$ therefore reveals the magnitude and mode of human influence, including whether continued intervention is required to maintain the state (Fig. 2c).

However, physical maintainability alone does not determine which balance domain should be pursued. A sustainable balance domain, $\beta_S$, is a dynamically maintainable region selected according to explicit environmental and societal values, including environmental safety, preservation of key functions, social priorities and acceptable maintenance requirements. Rather than arising from another boundary regime, $\beta_S$ is a normative selection from the dynamically maintainable state space. It identifies the target that environmental governance and intervention seek to make accessible and sustain (Fig. 2d).

Taken together, $\beta_0$, $\beta_N$ and $\beta_H$ describe how the maintainable state space differs under boundary-closed, naturally open and human-conditioned conditions, while $\beta_S$ identifies the sustainable domain to be pursued according to explicit environmental and societal values. Because all are represented within the same matter–energy state space and organized by EFE, their location, extent, relative stability, gradients and sustaining fluxes can in principle be learned, calculated and compared. The contributions of intrinsic dynamics, natural exchange and human action, as well as possible pathways towards $\beta_S$, can therefore be theoretically defined and potentially computed rather than remaining descriptive.

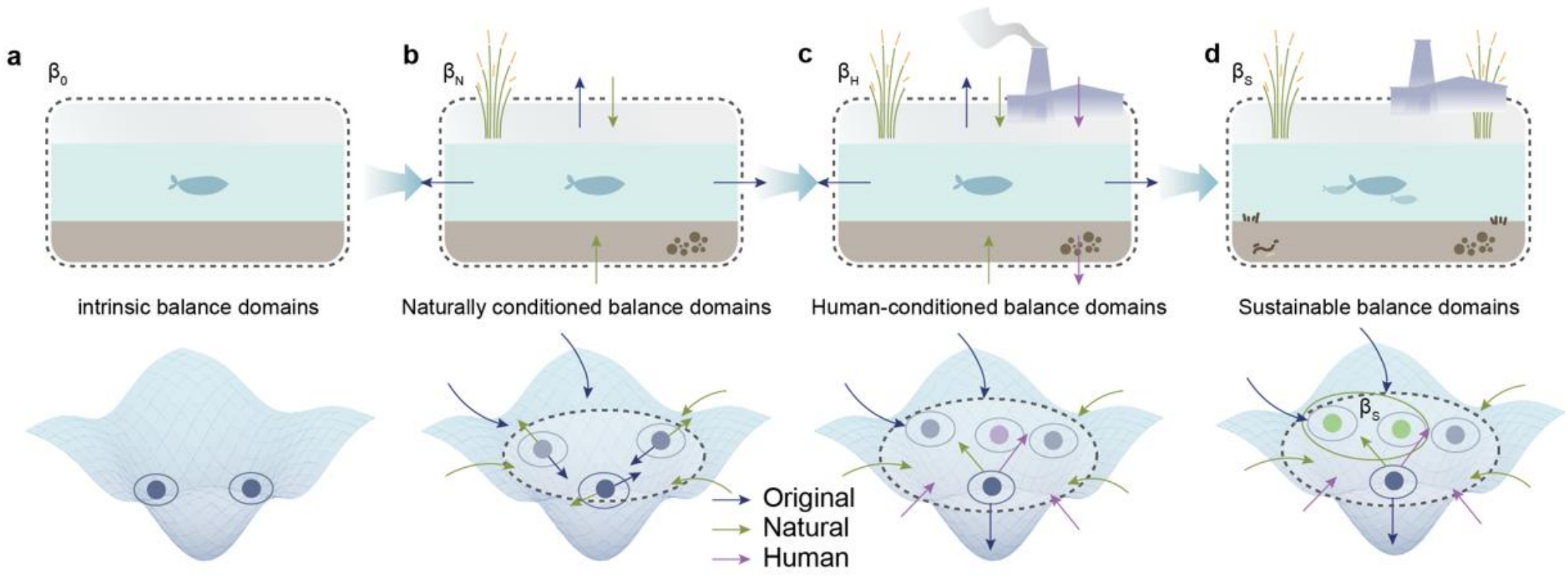

**Fig. 2 | Boundary-conditioned balance domains of the same system.** **a**, Boundary closure leaves only intrinsic dynamics, defining intrinsic balance domains $\beta_0$. **b**, Natural exchange reshapes the maintainable state space, producing naturally conditioned domains $\beta_N$. **c**, Human influence further modifies this state space, producing human-conditioned domains $\beta_H$. **d**, The sustainable balance domain $\beta_S$ is selected from dynamically maintainable regions according to environmental and societal values. Blue, green and purple arrows denote intrinsic dynamics, natural exchange and human influence, respectively.

## Making environmental balance computable

A framework becomes computationally useful when its objects and relations can be estimated from observations and tested against system behavior. This does not require a single closed-form equation for every environmental system. Thermodynamics, for example, provides common state variables, potentials and physical constraints, while the specific relations among them differ between specific systems[25]. The environmental balance framework follows an analogous division. State variables $X$, EFE, boundary contributions and balance domains provide a common theoretical structure, while the functions connecting them are specific to the environmental system and scale being studied. Constructing this structure draws on a series of supporting theories: thermodynamics and non-equilibrium landscape theory for potentials and physical constraints, dynamical-systems and stability theory for balance domains and transitions, and multiscale analysis for linking fine- and coarse-scale states. Each must be adapted to, and tested for, open environmental systems.

Learning these system-specific functions requires a tractable representation of the dynamics. H1–H4 assume conditional attraction to balance domains, scale-dependent continuity, effective stochasticity and approximate Markov closure. Together, they define the continuous stochastic state evolution to be learned from finite observations (Box 1).

| Hypothesis | Core statement | Interpretation and implications |
|---|---|---|
| **H1** | Conditional attraction to balance domains. | A system may be observed while moving within a transient pathway or between domains. If its boundary conditions and external driving persist for longer than its characteristic adjustment time, its trajectory will become statistically confined within one of the accessible balance domains. |
| **H2** | Scale-dependent continuity. | Environmental states evolve continuously at the selected physical scale. Rapid and critical transitions represent fast movement through state space rather than discontinuities in time. |
| **H3** | Effective stochasticity. | $X$ cannot resolve every microscopic process, local heterogeneity or fluctuating influence. Environmental trajectories and transitions between domains are therefore probabilistic at the modelling scale. |
| **H4** | Effective Markov closure. | When $X$ and the relevant boundary conditions sufficiently represent the information governing change, future evolution depends mainly on the current effective state rather than the complete history. Hidden variables, delayed responses or a finite window of recent observations may be incorporated when necessary, and the adequacy of this closure can be tested from data. |

**Box 1 | Hypotheses defining the learnable dynamics of environmental balance**

The relations governing this continuous stochastic state evolution can draw on several forms of physical and mechanistic knowledge. Conservation of matter and energy can be imposed directly, while established transport, reaction, biological and ecological relations can be incorporated wherever available[26–29]. For open, gradient-driven systems, non-equilibrium thermodynamics provides further structures by linking persistent gradients to fluxes and dissipation. Onsager-type force–flux and variational formulations characterize dissipative responses, while broader reversible–irreversible decompositions and stochastic thermodynamics accommodate non-dissipative motion, effective fluctuations and entropy production[30–32]. Together, these principles define the physically admissible structure of the dynamics while leaving its system-specific form to be learned from data.

However, defining the admissible structure does not determine the dynamics of a particular system. The remaining functions may couple many state variables, mechanisms and spatial scales, and are usually only partially specified by existing process knowledge. Their identification therefore requires evidence from system behaviors (Fig. 3a). Top-down Earth-system observation, from long-term records to satellite remote sensing of climatic variables, land cover, atmospheric composition and the surface ocean, delivers repeated, internally consistent coverage of environmental states across scales [33]. Bottom-up networks, such as autonomous ocean profiling floats and eddy-covariance flux towers, resolve the local exchanges of matter and energy that generate those large-scale patterns, while coordinated field campaigns link the two levels[34]. These observation sources provide such evidence for system-specific behaviors[35].

Structure-preserving AI can learn the unknown relations from these data while enforcing the theoretical structure and physical constraints defined above[36–39]. Typically, physics-guided neural networks have been widely applied in Earth and environmental sciences, embedding governing equations and conservation laws directly into the learning objective[40,41]. In environmental science, physics-guided learning have been used to reconstruct hydrodynamic fields from sparse observations[42], and to predict water resources[43] under physical constraints. It is common that AI turns the theoretical structure into a system-specific computational model without replacing its physical basis (Fig. 3b).

Once these relations are learned, both environmental states and pathways become calculable (Fig. 3c). The model can map balance domains, estimate their boundaries and relative stability, and calculate EFE gradients, sustaining fluxes, mismatch accumulation and rates of entropy production. It can further predict how the system will evolve under changing natural exchange or human influence and identify feasible routes towards a selected sustainable domain. Environmental analysis can therefore move from describing the present state to explaining how it is maintained and determining where the system can go.

These calculations are initially tied to the scale at which $X$ is defined, whereas the mechanisms maintaining environmental balance often operate across scales. A further computational task is therefore to connect fine-scale state coordinates with effective coarse-scale variables (Fig. 3d). The same structure-preserving learning can estimate how collective fine-scale dynamics generate coarse-scale EFE landscapes, balance domains and transition paths[44]. This may allow information to be aggregated across levels of organization and macroscopic environmental behavior to be related back to its underlying mechanisms.

The environmental balance framework could therefore become computable through the combination of common theoretical objects, physical constraints, environmental data and structure-preserving AI. The supporting theories specify what must be represented, physical principles constrain the admissible dynamics, and AI learns the system- and scale-specific relations from data. Together, these elements convert observations into estimates of which states can persist, what sustains them and which paths connect them, providing a computational basis for diagnosing existing balances and evaluating routes towards sustainable ones.

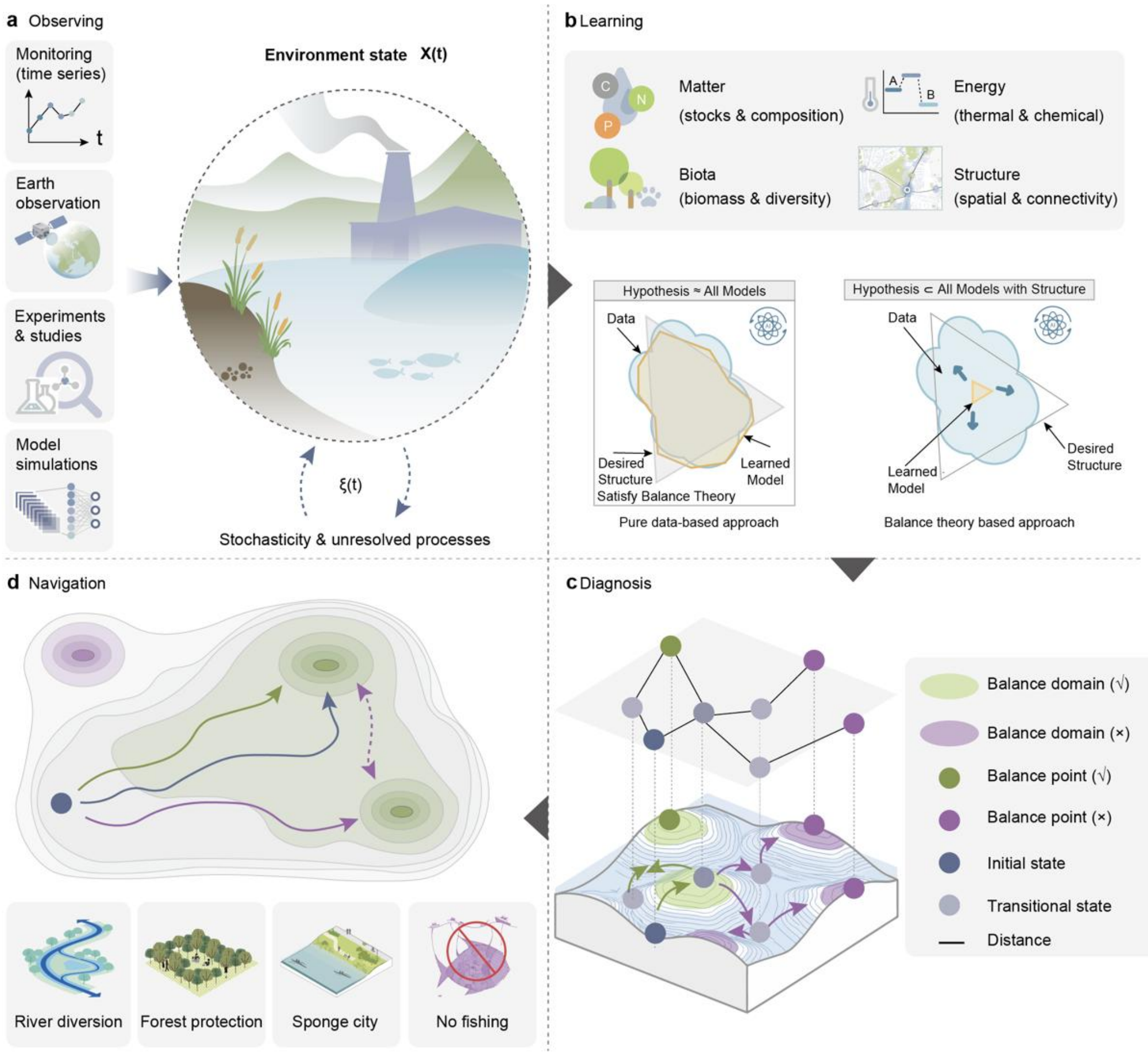

**Fig. 3 | Environmental balance computational framework.** **a**, Observations and simulations reconstruct environmental trajectories $X(t)$, $\xi(t)$ denotes unresolved variability. **b**, Matter–energy states and structural information are integrated to learn physically constrained system dynamics. **c**, The learned dynamics identify balance domains, their relative stability and possible transitions on the EFE landscape. Green and purple denote candidates that satisfy or fail the environmental and societal criteria, respectively. **d**, Alternative interventions are evaluated as pathways towards selected sustainable domains. EFE, environmental free energy.

## From environmental thresholds to balance navigation

The ability to calculate balance domains and transition paths could help close an important gap between setting environmental targets and determining how they can be achieved. Existing frameworks can define safe or sustainable conditions, assess whether a system can withstand disturbance and guide collective actions. A target condition alone, however, does not show which states consistent with it can persist under prevailing natural exchange and human influence, or which interventions can move the system into and maintain it within those states. The environmental balance framework complements these existing frameworks by providing a computable state–path approach for answering these questions (Box 2).

| Environmental framework | Primary question | Complement provided by the environmental balance framework |
|---|---|---|
| Environmental capacity and critical-load frameworks | How much external loading can a system receive before environmental standards or functions are compromised? | The framework examines the multidimensional states generated jointly by matter and energy inputs, internal transformations and ecological feedbacks, rather than treating system response only through individual loads. |
| Material-flow analysis and circular-economy frameworks | How do materials move through a system, and how can resource use, waste generation and environmental discharge be reduced? | The framework relates these flows to the states and functions they sustain, revealing how changes in material circulation may shift balance domains or open and close transition paths. |
| Resilience, multistability and critical-transition theory | Can a system withstand disturbance, recover or shift into an alternative regime? | The framework identifies the matter–energy relations sustaining different domains and represents change through accumulating mismatch and explicit pathways. It thereby addresses gradual drift and long transients as well as abrupt transitions. |
| Planetary boundaries, safe-and-just Earth-system boundaries and related sustainability frameworks | Which environmental conditions should society avoid or seek to remain within? | The framework examines which states consistent with those conditions are dynamically maintainable and how they can be reached and sustained under changing natural and human influence. |
| Social–ecological systems and adaptive-governance frameworks | How do environmental processes, human actors and institutions interact, and how should decisions adapt to change and uncertainty? | The framework represents how human actions alter environmental states, governing relations and accessible paths, allowing interventions to be evaluated through the balance domains they create or remove, the paths along which the system evolves and the continuing support required. |

**Box 2 | The environmental balance framework as a complement to established environmental frameworks**

The wider significance of this framework is threefold (Fig. 4). First, comparing the conditional domains defined above could make the relationship between humanity and nature theoretically explicit and potentially computable. It could reveal whether human activity shifts or contracts naturally maintainable domains, weakens the system's capacity to recover or establishes new balances dependent on continuing matter–energy inputs. Human influence would then be understood not merely as an added pollutant load, but as a systemic reorganization of state-generating mechanisms, feasible domains and transition paths. Harmony between humanity and nature could accordingly be assessed by whether this reorganization enables the selected sustainable domain to persist without accumulating mismatch or requiring unacceptable levels of continuing support.

Second, system-specific balance models could support more precise environmental management. They could reveal not only whether a target has been met, but how close the system is to leaving its present domain, what sustains that domain and where mismatch is accumulating. Interventions could then be chosen according to whether the state, boundary exchange or governing relations need to change, while candidate paths could be compared by their transition risks and maintenance demands. Environmental management would thereby move from one-off threshold compliance towards the continuing navigation of environmental states.

Third, cross-scale learning could allow environmental goals to define future technology requirements. Starting from a selected sustainable domain, learned cross-scale relations could translate the system-level changes required to reach and maintain it into finer-scale functional requirements. These requirements could guide the development of materials[45,46], catalysts[47,48], biological processes[49], sensing systems[50,51] and adaptive controls[52] with the specific properties needed to alter key fluxes, transformations or feedbacks. The framework could therefore connect a macroscopic environmental objective to the finer-scale technological capabilities required to realize it, rather than limiting analysis to technologies that already exist.

Taken together, these possibilities would transform a sustainable environmental condition from a static target into a domain whose generation, accessibility and maintenance can be investigated. Environmental thresholds would remain essential constraints, while balance navigation would address how a desirable state can be reached and sustained.

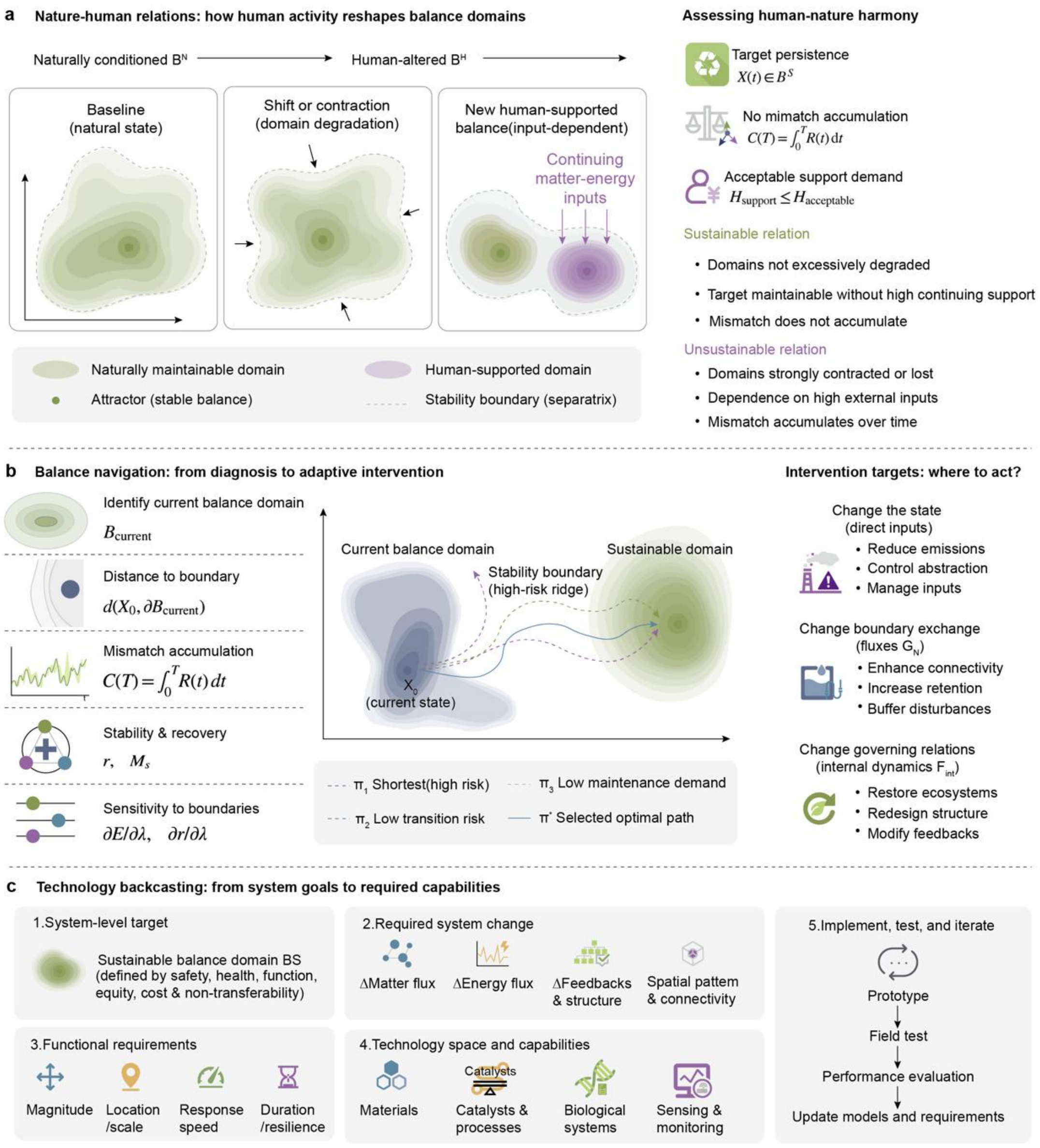

**Fig. 4 | The environmental balance framework as a state–path basis for assessment, intervention and innovation.** **a**, Reframing human–nature relations through changes in balance domains. **b**, Navigating from current to sustainable domains through model-guided intervention. **c**, Backcasting sustainable system goals into required technological capabilities.

## Outlook

Realizing the environmental balance framework will require a series of supporting theories and a staged research programme to develop and test them. The first priority is to formalize the mathematical relations among matter–energy states, EFE, intrinsic dynamics, boundary contributions, balance domains and transition paths. H1–H4 provide the simplifying structure needed to make this evolution tractable. Applications will then require appropriate variables, boundaries and scales, estimable quantities and system-specific dynamical closures. The resulting models can be evaluated by their ability to recover observed balance domains and predict trajectories under changing conditions. Where they fail, the mismatch would indicate which of H1–H4, or which supporting theory, needs revision.

The next priority is to combine data from environmental monitoring, Earth observation, experiments and process-based simulations with structure-preserving AI. These capabilities can be assembled into open and modular software for constructing state representations, learning system-specific dynamics, mapping balance domains, quantifying natural and human contributions and evaluating pathways towards sustainable domains. Shared datasets, benchmark systems and uncertainty assessment will support comparison and extension across systems and scales.

A third priority is to connect these objects to the macro-scale environmental agenda that motivates them. Planetary boundaries, the Paris Agreement temperature goal and water quality and water resource targets are stated as thresholds and target levels. The framework can restate them as questions about maintainable domains. On this basis, periodic assessment cycles could report not only distance to a threshold, but also the feasibility, transition risk and maintenance cost of the pathways that remain. Climate and water objectives could then be represented in one state and path structure rather than assessed separately.

Once mature, the environmental balance framework, together with its supporting theories, could provide a physically grounded basis for environmental science by connecting observed change to the matter–energy relations that generate and sustain it. It could enable quantitative understanding of human–nature relations, support more precise environmental management and translate system-level objectives into technological requirements across scales. At a more distant frontier, the same principles could guide the construction of artificial ecosystems beyond Earth, where material stocks, energy flows, boundary exchange and biological feedbacks must be deliberately organized to maintain habitable environmental domains.